\documentclass[journal,twoside,web]{ieeecolor2}
\usepackage{generic}
\usepackage{cite}
\usepackage{amsmath,amssymb,amsfonts}
\usepackage{algorithmic}
\usepackage{graphicx}
\usepackage{multirow}
\usepackage{hyperref} 
\usepackage{algorithm}
 \usepackage{booktabs} 
 \usepackage{tabularray}
\usepackage{float}

\def\BibTeX{{\rm B\kern-.05em{\sc i\kern-.025em b}\kern-.08em
    T\kern-.1667em\lower.7ex\hbox{E}\kern-.125emX}}
\begin{document}
\title{Addressing data annotation scarcity in Brain Tumor Segmentation on 3D MRI scan Using a Semi-Supervised Teacher-Student Framework}

\author{
    Jiaming Liu, Cheng Ding$^{*}$,and Daoqiang Zhang 
    \thanks{All authors are with the College of Artificial Intelligence, Nanjing University of Aeronautics and Astronautics, Key Laboratory of Brain-Machine Intelligence Technology, Ministry of Education, Nanjing 211106, China. (e-mail: gaming@nuaa.edu.cn;  dqzhang@nuaa.edu.cn;
    chengding@nuaa.edu.cn)}%
    \thanks{Daoqiang Zhang is also with the Shenzhen Research Institute, Nanjing University of Aeronautics and Astronautics, Shenzhen 518000, China.}%
    \thanks{*Corresponding author: Cheng Ding (chengding@nuaa.edu.cn).}
    \thanks{This work was supported in part by the National Natural Science Foundation of China under Grants 62136004 and 62276130, in part by the National Key R\&D Program of China under Grant 2023YFF1204803, and in part by the Key Research and Development Plan of Jiangsu Province under Grant BE2022842.}}

\maketitle

\begin{abstract}
Accurate brain–tumor segmentation from MRI is limited by expensive annotations and data heterogeneity across scanners and sites. We propose a semi-supervised teacher–student framework that combines an uncertainty-aware pseudo-labeling teacher with a progressive, confidence-based curriculum for the student. The teacher produces probabilistic masks and per-pixel uncertainty; unlabeled scans are ranked by image-level confidence and introduced in stages, while a dual-loss objective trains the student to learn from high-confidence regions and unlearn low-confidence ones. Agreement-based refinement further improves pseudo-label quality. On BraTS 2021, validation DSC increased from 0.393 (10\% data) to 0.872 (100\%), with the largest gains in early stages, demonstrating data efficiency. The teacher reached a validation DSC of 0.922, and the student surpassed the teacher on tumor subregions (e.g., NCR/NET 0.797 and Edema 0.980); notably, the student recovered the Enhancing class (DSC 0.620) where the teacher failed. These results show that confidence-driven curricula and selective unlearning provide robust segmentation under limited supervision and noisy pseudo-labels.
\end{abstract}


\section{INTRODUCTION}

Brain tumor segmentation from magnetic resonance imaging (MRI) plays a vital role in clinical practice  \cite{hosny2018artificial,esteva2019guide}. Accurate segmentation supports diagnosis, treatment planning, surgical guidance, and monitoring of disease progression. In recent years, deep learning models, especially UNet-based architectures, have achieved notable success in medical image segmentation tasks \cite{ronneberger2015unet}. However, these models rely heavily on large-scale annotated datasets. In medical imaging, acquiring such datasets is challenging because expert radiologists must provide detailed pixel-level annotations, which is both time-consuming and costly. Moreover, one of the main challenges in brain tumor segmentation is the heterogeneity of MRI data \cite{karun2025hybrid}. Models trained on small and uniform datasets often might not perform well when applied to real clinical data with such variability.

\par To address these challenges, semi-supervised learning (SSL) has been introduced, combining limited labeled data with large amount of unlabeled data \cite{bai2017semi}. This makes it possible to adapt models to diverse patient populations, imaging devices, and clinical settings, improving their robustness in real-world practice. Beyond reducing annotation costs, SSL also promotes faster development of decision-support tools, shortens the time to clinical translation, and ensures that segmentation models remain effective as medical data grows \cite{huang2023selfsupervised_review}. For brain tumor analysis, SSL therefore can act as a bridge between research settings with carefully curated datasets and the heterogeneous data encountered in hospitals, making advanced AI tools more practical and impactful for routine clinical workflows \cite{wang2023ssl}.

Despite these improvements, important challenges remain. First, pseudo-labels generated from unlabeled scans often contain errors, especially in low-confidence tumor regions, which can negatively affect learning \cite{yu2019uncertainty}. Second, most approaches select pseudo-labeled data only once, without progressively updating them during training. Third, many models produce only segmentation masks without estimating uncertainty, which limits their reliability in clinical use. Addressing these issues is crucial for building robust semi-supervised methods that can operate reliably in clinical environments.

\par In this study, we propose a semi-supervised teacher–student framework introducing a novel TransASPP-UNet architecture to overcome the problem of annotation scarcity and data heterogeneity. The teacher model generates probabilistic pseudo-labels along with per-pixel uncertainty maps, allowing unlabeled samples to be ranked by confidence. The student model begins training with the top m\% of high-confidence samples, and progressively incorporates more samples in later stages. To improve robustness, the student is trained with a dual-loss objective that minimizes loss on high-confidence samples and maximizes loss on low-confidence samples, thereby “unlearning” unreliable information. Furthermore, a feedback loop compares teacher and student predictions: regions of agreement are retained, while uncertain areas are refined, leading to better pseudo-label quality over time. The main contributions of this work are summarized as follows:
1)	We design a teacher–student semi-supervised framework that integrates uncertainty-aware pseudo-labeling with a progressive confidence-based sampling strategy, where the student model gradually expands training with more unlabeled data.
2)	We introduce a dual-loss function that incorporates both supervised learning and unlearning from low-confidence predictions.
3)	In this study, a feedback refinement mechanism is proposed based on teacher–student agreement to improve pseudo-label quality and develop a novel TransASPP-UNet architecture, which integrates attention gates, squeeze-and-excitation blocks, ASPP, and dual output heads for segmentation and uncertainty estimation.
Together, these contributions enable more effective use of unlabeled MRI data and provide a practical solution for brain tumor segmentation under annotation scarcity.

\section{Literature Review}

Deep learning--based brain tumor segmentation on MRI scans has achieved remarkable success, yet its dependency on large-scale annotated data remains a major bottleneck. To address this, SSL strategies have been extensively studied. 
Tarvainen et al. introduced the Mean Teacher framework, where a student network learns from an exponential moving average (EMA) teacher under perturbations \cite{tarvainen2017mean}. Building on this idea, Li et al.\ proposed SASSNet, combining segmentation with signed distance maps to embed shape priors and enhance tumor boundary delineation \cite{li2020sassnet}. Chen et al.\ developed multi-scale mean-teacher variants with adversarial regularization to improve generalization for lesion segmentation \cite{chen2021adversarialmt}. Xu et al.\ later introduced ambiguity-selective consistency, directing the model toward uncertain regions such as tumor boundaries \cite{xu2023ambiguity}. Similarly, Lu et al.\ proposed uncertainty-aware pseudo-labeling, in which pseudo-labels are filtered by voxel-level confidence \cite{lu2023uncertainty}.  
Chen et al.\ proposed Cross Pseudo Supervision (CPS), in which two models generate and supervise each other’s pseudo-labels \cite{Chen2021CPS}. Wu et al.\ extended this idea with MC-Net+, a mutual consistency network with multiple decoders that reinforces representation learning under label scarcity \cite{wu2022mcnet}. Liu et al.\ further advanced pseudo-labeling with soft probabilistic masks to reduce confirmation bias in heterogeneous tumor regions \cite{Liu2024MOST}.  
Zhu et al.\ introduced a Hybrid Dual Mean-Teacher Network with double-uncertainty guidance, demonstrating improved tumor segmentation under limited annotations \cite{zhu2024dualmt}. He et al.\ \cite{He2025EPL} proposed Evidential Prototype Learning (EPL), which fuses evidential predictions with dual-uncertainty masking to enhance voxel-wise semi-supervised segmentation in a unified framework. More recently, Assefa et al.\ presented DyCON, which integrates dynamic uncertainty-aware consistency with contrastive learning, further stabilizing pseudo-label training across modalities \cite{assefa2025dycon}. Karri et al.\ designed UG-CEMT, an uncertainty-guided cross-attention ensemble mean teacher using Vision Transformers, which approaches fully supervised performance with only 10\% labeled data \cite{karri2024ugcmt}.  
In MRI brain tumor segmentation,  Devi et al.\ proposed motion-artifact-augmented pseudo-labeling to increase robustness against acquisition noise \cite{Qu_2024}. Ottesen et al.\ recently validated SSL clinically by demonstrating its effectiveness in brain metastasis segmentation under real-world annotation scarcity \cite{ottesen2025ssl}.  

Despite these advances, several gaps remain. First, low-confidence predictions are typically ignored or softly down-weighted, but explicit unlearning strategies are underexplored. Finally, while agreement filtering between models has been studied, adaptive feedback loops in which student performance dynamically adjusts teacher sampling strategies are uncommon.

\section{Methodology}
\begin{figure*}[!t]
    \centering
    \includegraphics[width=0.75\textwidth]{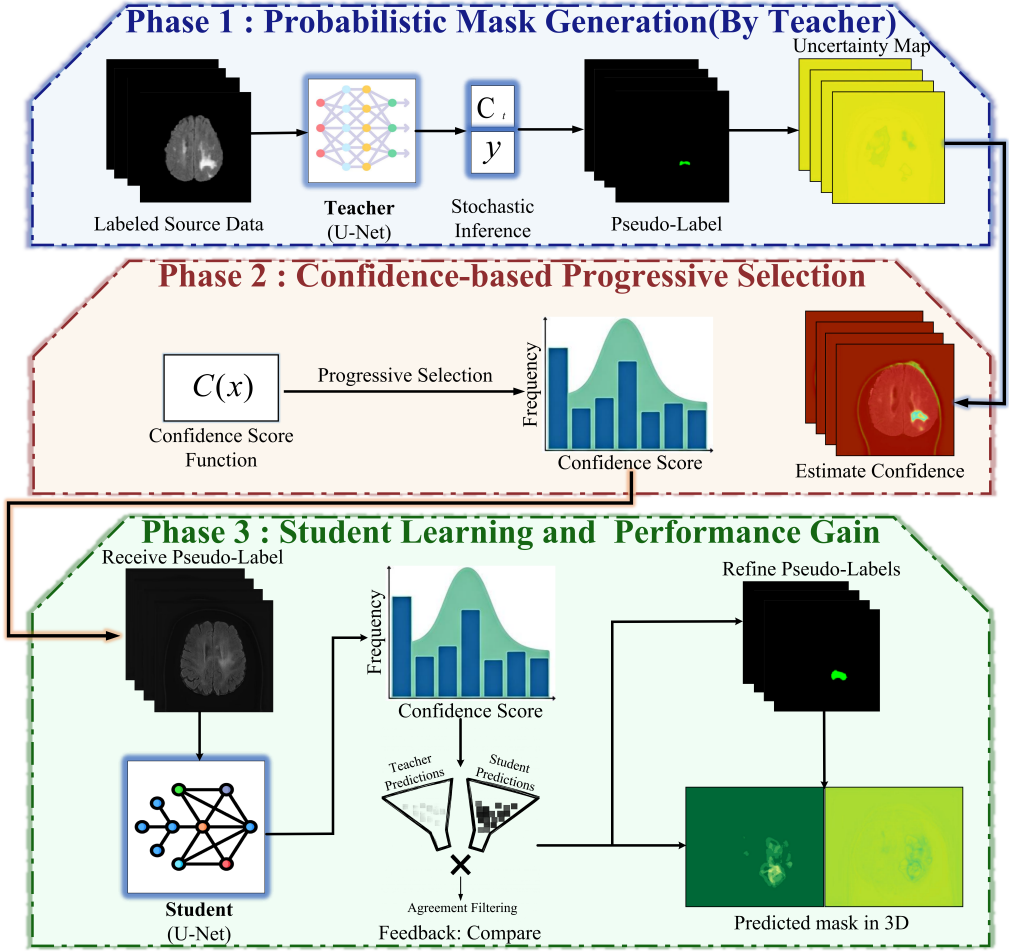}
    \caption{Progressive pseudo-labeling framework for semi-supervised brain tumor segmentation. The Teacher U-Net is first trained on labeled MRI data to generate probabilistic segmentation masks and uncertainty estimates. Pseudo-labels are then generated for unlabeled target-domain images and ranked by confidence scores . A progressive sampling strategy selects high-confidence samples to train the Student model in stages, with feedback refinement comparing teacher-student predictions to improve pseudo-label quality iteratively.}
    \label{fig:progressive_pseudolabeling_framework}
\end{figure*}

An overview of the complete pipeline of this study is presented in Figure \ref{fig:progressive_pseudolabeling_framework}.

\par This framework begins by training a Teacher U-Net model on labeled source-domain MRI data using a joint Dice and Cross-Entropy loss. The trained Teacher then generates pseudo-labels (probabilistic masks) for unlabeled target-domain MRI images. Confidence scores are estimated for each sample and used to rank pseudo-labeled images. A progressive sampling strategy selects the top-M\% most confident samples to train a student model. This iterative process refines segmentation performance across domains using both labeled and unlabeled data.

\subsection{Dataset Description} 
\begin{figure}[H] 
    \centering
    \includegraphics[width=\columnwidth]{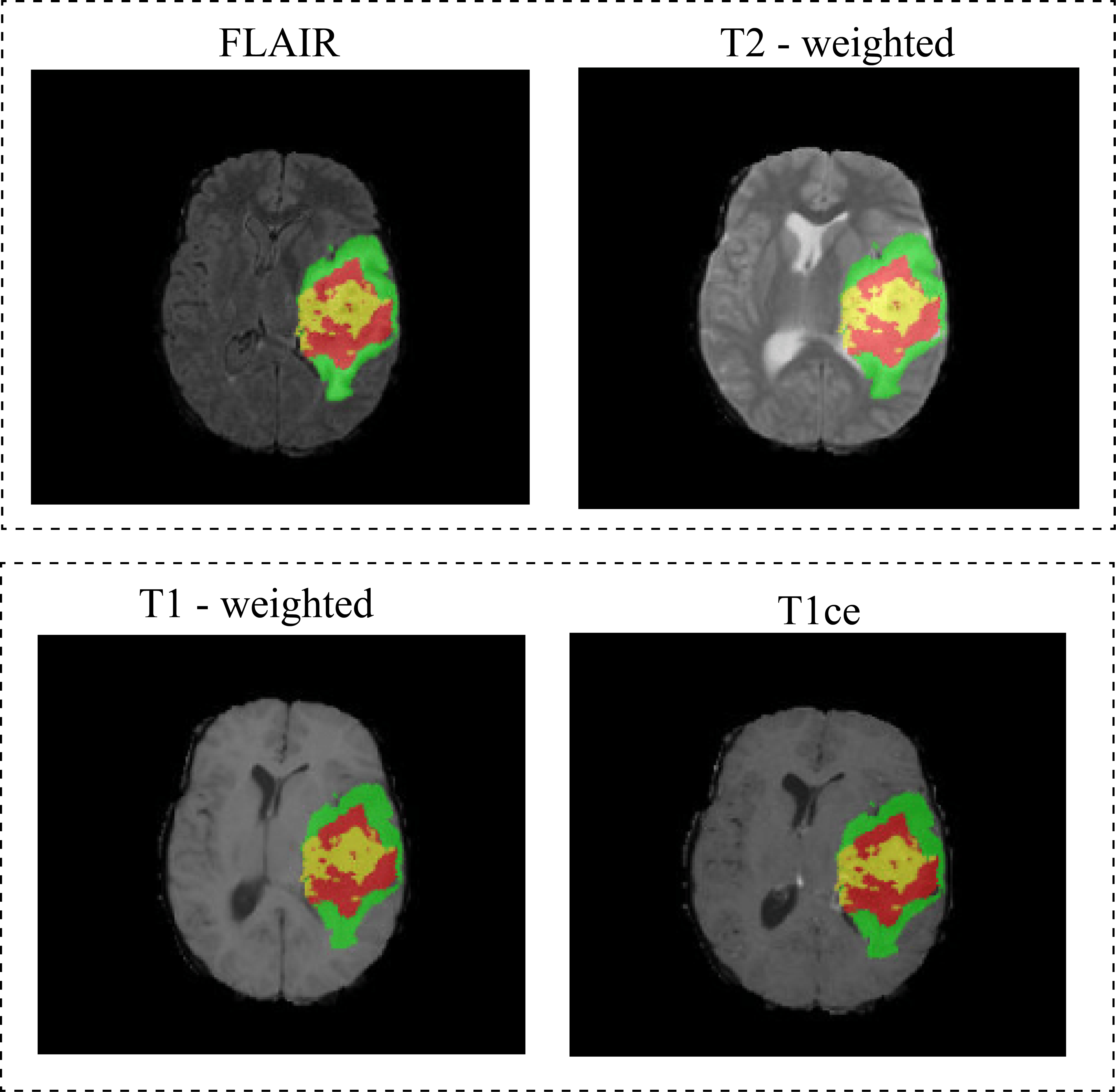}
    \caption{Overlay of tumor subregions (enhancing tumor, tumor core, and edema) on T1, T1ce, T2, and FLAIR MRI sequences.}
    \label{fig:glioma_tumor_subregions_mri}
\end{figure}

We have used the BraTS 2021 dataset \cite{menze2015brats}, which was released as part of the MICCAI BraTS 2021 challenge. It includes multi-institutional and multi-parametric MRI scans of glioma patients, collected and standardized across different scanners and clinical sites. An illustration of brain sets is shown in Figure \ref{fig:glioma_tumor_subregions_mri}.

\par Each patient case contains four MRI modalities: native T1-weighted (T1), post-contrast T1-weighted (T1ce), T2-weighted (T2), and T2 Fluid Attenuated Inversion Recovery (T2-FLAIR). These modalities provide complementary structural and pathological information. T1 and T1ce capture fine anatomical details and highlight enhancing tumor regions, T2 highlights tumor-associated edema, while T2-FLAIR suppresses fluid signals, making peritumoral edema more visible. The BraTS 2021 dataset contains a total of 2,000 cases. The dataset is divided into training, validation, and test sets. The training set includes 1,251 real cases, with all four MRI modalities and ground truth annotations provided. The validation set includes 219 cases, and the test set consists of 530 cases.

\subsection{Proposed Approach} 
We formulate the brain tumor segmentation problem under a semi-supervised teacher--student setting. 
Let the labeled dataset be denoted as:

\begin{equation}
S = \{(x_i, y_i)\}_{i=1}^{N_\ell},
\label{eq:1}
\end{equation}

where each \( x_i \in \mathbb{R}^{H \times W} \) represents an MRI slice of spatial size \( H \times W \), and \( y_i \in \{0,1\}^C \) is the corresponding one-hot voxel-level segmentation mask for \( C \) tumor classes. The unlabeled dataset is denoted by \( U = \{x_j\}_{j=1}^{N_u} \), which contains images without annotations. 
Two models are employed the teacher network \( M_T(\cdot; \theta_T) \) with parameters \( \theta_T \), and the student network \( M_S(\cdot; \theta_S) \) with parameters \( \theta_S \). For any input image \( x \), both networks output a segmentation logit map \cite{Lakshmi2025} \( Z(x) \in \mathbb{R}^{H \times W \times C} \) and a corresponding softmax probability distribution as equation \ref{eq:2}, where \( p \in \{1,\dots,H \times W\} \).  
\begin{equation}
P(x)_{p,c} = \text{softmax}_c (Z(x)_{p,c}), 
\label{eq:2}
\end{equation}

In addition, each network produces an uncertainty map \cite{Emon2025} \( U(x)_p = \log \sigma^2(x)_p \), which represents the per-pixel predictive variance. For compact notation, we write teacher outputs as \( (P_T, U_T) = M_T(x; \theta_T) \) and student outputs as \( (P_S, U_S) = M_S(x; \theta_S) \). This formalism sets the foundation for supervised learning on \( S \), pseudo-label generation from \( U \), and uncertainty-aware consistency between teacher and student.

\subsubsection{TransASPP-UNet Backbone} 
\begin{figure*}[!t]  
    \centering
    \includegraphics[width=\textwidth]{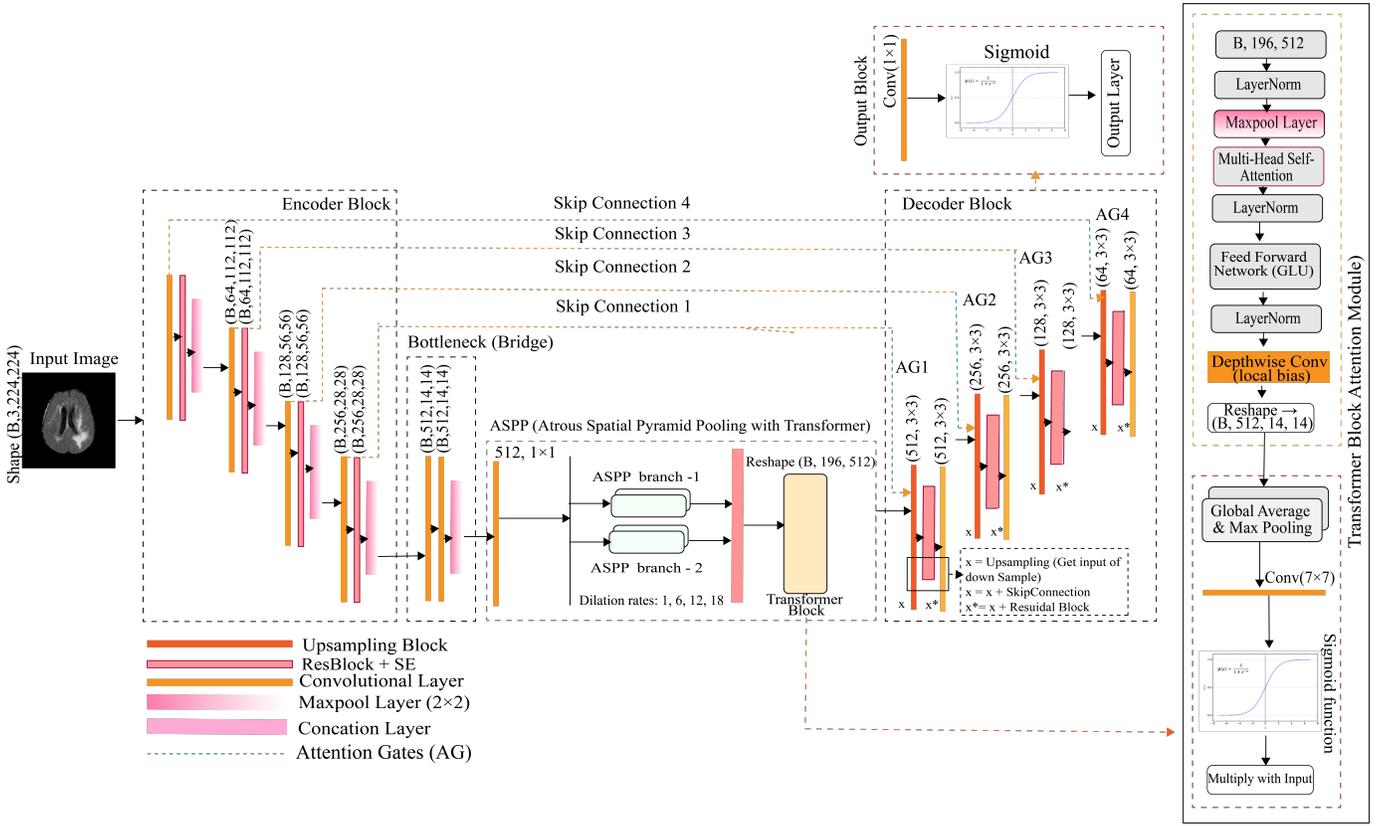}
    \caption{Schematic of the proposed U-Net–based segmentation model integrating Atrous Spatial Pyramid Pooling (ASPP) and attention transformer to enhance multiscale feature learning and focus on salient regions.}
    \label{fig:transaspp_module_architecture}
\end{figure*}

The proposed TransASPP-UNet is a hybrid encoder–decoder segmentation model that integrates Atrous Spatial Pyramid Pooling (ASPP) \cite{Wang2024} with a customized Transformer block to effectively capture both local and global contextual features. The architecture consists of four main components: the encoder, the ASPP module, the transformer-enhanced bottleneck \cite{HANINE2024109440}, and the decoder. The proposed TransASPP-UNet architecture is demonstrated in Figure 
\ref{fig:transaspp_module_architecture}.

\textbf{Encoder Path.} The encoder is composed of convolutional layers, residual blocks, and max-pooling layers that progressively reduce the spatial resolution while increasing the feature depth. Given an input image 
$X \in \mathbb{R}^{B \times H \times W \times C}$, where $B$ is the batch size, $H$ and $W$ are the height and width, and $C$ is the number of channels, the feature extraction at each stage is defined as:
\begin{equation}
F_{\text{enc}}^{(l)} = \sigma \left( W^{(l)} * F_{\text{enc}}^{(l-1)} + b^{(l)} \right),
\end{equation}
where $*$ denotes convolution, $\sigma$ is the non-linear activation function (ReLU) \cite{Panigrahi2024}, and $F_{\text{enc}}^{(0)} = X$. After each stage, max pooling reduces the spatial resolution. This encoder generates a hierarchy of feature maps that are later passed to the decoder via skip connections.

\textbf{Atrous Spatial Pyramid Pooling (ASPP).}
To capture multi-scale contextual information, the deepest encoder feature map is passed through the ASPP module, which applies multiple atrous convolutions with different dilation rates $\{r_1, r_2, \dots, r_n\}$ \cite{LIU2024102776}.

This ensures that both fine-grained details and long-range dependencies are retained.

\textbf{Transformer Block (Bottleneck with Global Attention).}
The ASPP features are reshaped into a sequence of tokens \cite{Wang2024ICBASE} as:
\begin{equation}
F_{\text{seq}} = \text{Reshape}(F_{\text{aspp}}) \in \mathbb{R}^{B \times N \times d},
\end{equation}
where $N$ is the number of tokens and $d$ is the feature dimension. The Transformer block then models long-range dependencies. The sequence features are projected into query, key, and value spaces. The attention mechanism computes similarity between queries and keys, weighting the values accordingly \cite{Wang2024ICBASE}. In terms of multi-head attention, for $h$ attention heads, the outputs are concatenated and projected. This enables the model to jointly attend to features from multiple subspaces \cite{Rasool2024}. Each token embedding is further refined with a two-layer feed-forward network \cite{AlHasan2025}. Later, to stabilize training and preserve information flow, residual connections are applied. Thus, the Transformer bottleneck provides globally contextualized features.

\textbf{Decoder Path.}
The decoder reconstructs the segmentation map by progressively upsampling feature maps and fusing them with encoder features via skip connections \cite{ZHU2024108284}. At each stage:
\begin{equation}
F_{\text{dec}}^{(l)} = \sigma \left(W_{\text{up}}^{(l)} * \text{Upsample}(F_{\text{dec}}^{(l+1)}) + W_{\text{skip}}^{(l)} * F_{\text{enc}}^{(l)} \right)
\end{equation}
where $W_{\text{up}}^{(l)}$ are learnable parameters for upsampling \cite{Tiwary2025}, and $W_{\text{skip}}^{(l)}$ integrates encoder features. Residual connections within decoder blocks further refine contextual details.

The final decoder output is passed through a $1 \times 1$ convolution followed by a sigmoid activation \cite{Hasan2024} to generate a binary segmentation mask.

\subsubsection{Teacher Training and Uncertainty Distillation} 
The teacher is first trained in a fully supervised manner on \(S\). Its objective combines Dice loss and cross-entropy loss, regularized by weight decay \cite{AN2024110731}:

\begin{equation}
\begin{aligned}
L_{\text{teacher}}(\theta_T) &= \sum_{(x,y)\in S} \Bigg[ L_{\text{Dice}}(P_T(x),y) \\
&\quad + \lambda_{\text{CE}} \, L_{\text{CE}}(P_T(x),y) \Bigg] \\
&\quad + \lambda_{\text{reg}} \lVert \theta_T \rVert^2  
\end{aligned}.
\end{equation}

Here, \(L_{\text{teacher}}(\theta_T)\) is the total supervised objective for the teacher model. The first two terms enforce accurate segmentation using Dice and cross-entropy losses, while the last term is an \(L_2\)-regularization penalty controlled by \(\lambda_{\text{reg}}\). 
$P_T(x)$ is the teacher’s voxel-wise class-probability map \cite{Guo2025}, $L_{\text{Dice}}$: = (soft) Dice loss, $L_{CE}$ = pixel-wise cross-entropy loss. This loss measures the negative log-likelihood of the predicted probability distribution \(P\) compared to the ground truth labels \(y\), averaged over all pixels \(p\) and classes \(c\). The soft Dice loss encourages spatial overlap between the prediction and ground truth, with \(\epsilon\) ensuring numerical stability \cite{Rasool2025Review}.  

To train the uncertainty head \cite{Hassan2024}, variance is approximated via \(K\) stochastic forward passes (using dropout or data augmentations). The target log-variance provides a stabilized training target for the uncertainty estimation head by mapping empirical variance into the log domain \cite{Yang2024HippocampusSegmentation}. The uncertainty regression loss then enforces the predicted uncertainty map \(U_T(x)\) to align with the target log-variance, averaged across both labeled and unlabeled data.

\subsubsection{Pseudo-Label Generation and Progressive Sampling}
For each unlabeled image \(x\), the teacher produces an averaged probability map
\(\bar P_T(x)=\tfrac{1}{K}\sum_{k=1}^{K}P_T^{(k)}(x)\), from which pseudo-labels
\(\hat y(x)\) are taken as hard \(\arg\max\) or soft probabilities \cite{Chatterjee2025PseudoLabel}.The per-pixel confidence score, denoted as \( c_p(x) \), represents the model’s degree of certainty or reliability regarding its prediction for each individual pixel \( x \) in an image. To measure reliability, a per-pixel confidence score \(c_p(x)\) is computed as which down-weights uncertain predictions using the teacher’s uncertainty map \(U_T(x)\) \cite{Zhang_2024_CVPR}.  An image-level confidence \cite{Ran2025PseudoLabelReview} is then defined as the mean pixel confidence:  
\begin{equation}
C(x) = \frac{1}{HW} \sum_{p} c_p(x) .
\end{equation}  

Unlabeled samples are ranked by \(C(x)\), and at iteration \(t\), only the top \(K_t\) fraction is included:  
\begin{equation}
U_t = \{ x \in U : \text{rank}(C(x)) \leq K_t \cdot N_u \} ,
\end{equation}  
where \(N_u = |U|\) is the number of unlabeled samples. This ensures that the student starts with highly reliable pseudo-labels and gradually incorporates harder, less confident cases.

\subsubsection{Student Training with Dual-Loss Objective}
At cycle \(t\), the student is trained on the mixed dataset:  \(T_t = S \cup U_t\) where \(S\) is the labeled dataset and \(U_t\) is the subset of pseudo-labeled samples selected at iteration \(t\) \cite{Zheng2025MTSNet}.  
For each pseudo-labeled image \(x\), pixels are partitioned into high-confidence and low-confidence sets:  

\begin{equation}
P_{\text{high}}(x) = \{ p : c_p(x) \geq \tau_p \} ,
\end{equation}  

\begin{equation}
P_{\text{low}}(x) = \{ p : c_p(x) < \tau_p \} .
\end{equation}

where \(P_{\text{high}}(x)\) denotes the set of pixels whose confidence is above the threshold \(\tau_p\) and \(P_{\text{low}}(x)\) represents the pixels with confidence below the threshold, considered unreliable.  
The student objective is a dual loss \cite{AlZubi2025Federated}:

\begin{equation}
\begin{aligned}
L_{\text{total}}(\theta_S) &= L_{\text{sup}}(\theta_S) + \lambda_1 L_{\text{high}}(\theta_S) \\
&\quad - \lambda_2 L_{\text{low}}(\theta_S) + \lambda_{\text{reg}} \|\theta_S\|^2 ,
\end{aligned}
\end{equation}

where \(L_{\text{sup}}\) is the supervised loss (same as the teacher’s loss), \(\lambda_1\) balances the high-confidence loss \cite{Ziaee2025}, \(\lambda_2\) penalizes low-confidence fitting, and \(\lambda_{\text{reg}}\) controls weight regularization.  

The high-confidence loss is defined as:  
\begin{equation}L_{\text{high}}(\theta_S) = \sum_{x \in U} \frac{1}{|P_{\text{high}}|} \sum_{p \in P_{\text{high}}} L_{\text{pixel}} \big( P_{S,p}(x) , \hat{y}_p \big) ,\end{equation}
which averages the pixel-wise cross-entropy loss \cite{Arora2024KidneySegmentation} over all high-confidence pixels, ensuring the student learns from trustworthy pseudo-labels.  

The low-confidence loss is given by:  
\begin{equation}L_{\text{low}}(\theta_S) = \sum_{x \in U} \frac{1}{|P_{\text{low}}|} \sum_{p \in P_{\text{low}}} L_{\text{pixel}} \big( P_{S,p}(x) , \hat{y}_p \big) ,\end{equation}
This term measures the cross-entropy over low-confidence pixels. By subtracting it in the objective, the student is discouraged from memorizing noisy or unreliable pseudo-labels.  

Here, \(L_{\text{pixel}}(\cdot, \cdot)\) is the standard pixel-wise cross-entropy between the prediction and the pseudo-label. Intuitively, \(\lambda_1 L_{\text{high}}\) enforces learning from reliable regions, while \(-\lambda_2 L_{\text{low}}\) prevents overfitting to noise \cite{AbouAli2025}. This dual strategy allows the student to simultaneously learn accurate tumor structures and ignore uncertain areas, thereby improving segmentation \cite{PANI2024109418} under limited annotations.

\section{Results}
\begin{figure*}[htbp]
  \centering
  \includegraphics[width=\linewidth]{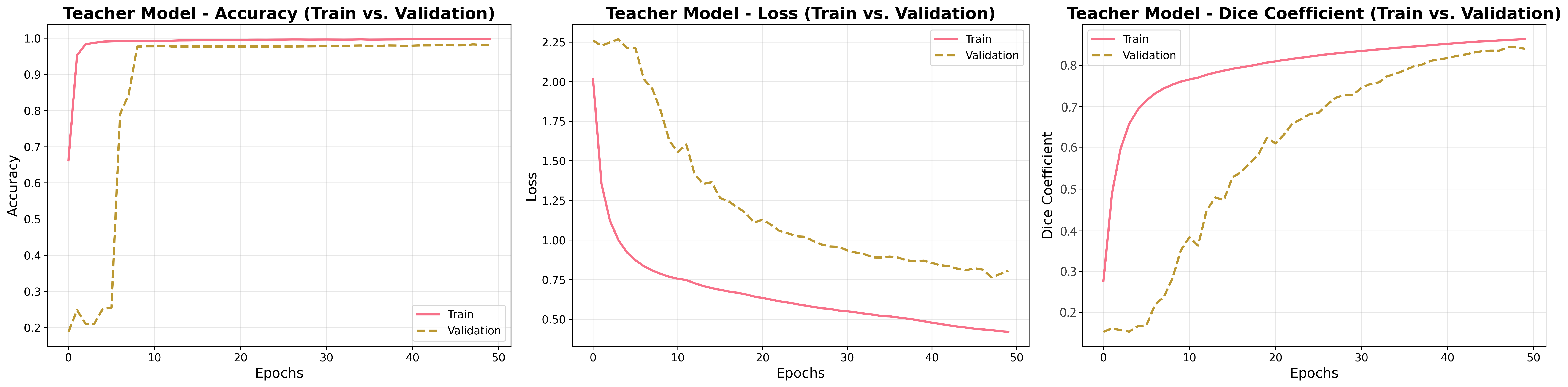}
  \caption{Learning Curves of the Teacher Model: Training vs. Validation Performance (Accuracy, Loss, Dice Coefficient over Epochs).}
  \label{fig:teacher_training} 
\end{figure*}

In this study, we comprehensively evaluated the segmentation performance of our proposed framework using various metrics, including the Dice coefficient (DSC), loss values, confidence analysis, and error analysis. We assessed the performance of both the teacher and student models to provide a holistic understanding of the system's effectiveness.
The training process began with the teacher model, which was trained on labeled MRI data to generate high-quality pseudo-labels for the unlabeled target data. Figure~\ref{fig:teacher_training} illustrates the training dynamics of the teacher model, showcasing the accuracy curve, loss curve, and Dice coefficient curve. These curves provide insights into the model's convergence behavior and its ability to generalize effectively to the segmentation task.

In Figure~\ref{fig:teacher_training} over the 50 epochs, the teacher model’s performance improves markedly. By around epoch 5‑10, training accuracy has already risen to ~0.98, eventually reaching nearly 1.00, while validation accuracy climbs more slowly, surpassing ~0.90 only after about 10 epochs, and approaching ~0.99 by the end. Similarly, training loss begins around ~2.00, falls rapidly in early epochs, and continues decreasing to about 0.45 by epoch 50. The validation loss also starts high (just above 2.00), declines more gradually with fluctuations, and ends around 0.80. The Dice coefficient for training starts near 0.35‑0.40, rises steeply in early epochs, reaching ~0.80 by mid‑training, and approaches ~0.82 or higher by epoch 50. The validation Dice coefficient starts very low (~0.10‑0.25), shows a steady increase (passing ~0.70 by epoch ~25), and by the final epochs attains ~0.82, nearly matching the training Dice. Together, these numbers suggest fast early learning, consistent improvement in validation metrics, and close convergence between training and validation by the end, indicating good generalization.

\begin{figure}[t!]
  \centering
  \includegraphics[width=0.5\textwidth]{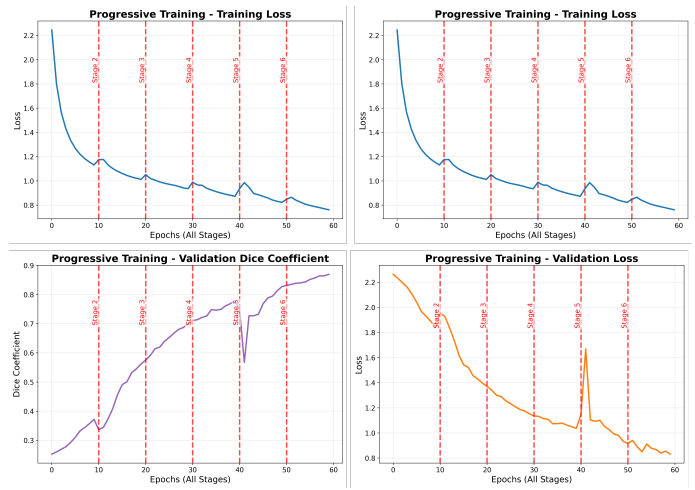}
  \caption{Progressive training dynamics across six curriculum stages (red dashed lines) over 60 epochs.  Training Dice (top-left) improves from ~0.3 to 0.87;  validation Dice (top-right) increases from 0.393 to 0.872 with temporary Stage 5 dip.  Training loss (bottom-left) decreases from ~2.2 to $<$0.8;  validation loss (bottom-right) declines from ~1.8 to ~0.88 with transient Stage 5 spike.  Curves demonstrate effective progressive learning with diminishing later-stage gains.}
  \label{fig:student}
\end{figure}

Further, we have evaluated the student model's performance during progressive training, a structured approach where the training process is divided into multiple stages. Each stage introduces new challenges or data complexities, allowing the model to gradually improve its learning capabilities. The figure \ref{fig:student} provides a comprehensive view of the model's performance metrics, including the Dice coefficient (accuracy) and loss (error), across six progressive training stages.

Figure\ref{fig:student}  illustrates the performance metrics of the student model during progressive training across six stages, with each stage separated by red dashed lines. The top-left plot shows the training Dice coefficient, which steadily improves from approximately 0.3 to 0.87 over 60 epochs, indicating enhanced model accuracy on training data. The top-right plot depicts the validation Dice coefficient, reflecting a similar trend, with accuracy on unseen data increasing consistently, except for a slight dip in stage 5. The bottom-left plot presents the training loss, which continuously decreases from around 2.2 to below 0.8, indicating reduced errors during training. Similarly, the bottom-right plot shows the validation loss, which also declines over time, with a temporary spike in stage 5, corresponding to the dip in the validation Dice coefficient. Overall, the plots demonstrate the model's progressive improvement in both training and validation performance across the six training stages.

\begin{figure}[t!]
  \centering
  \includegraphics[width=\linewidth]{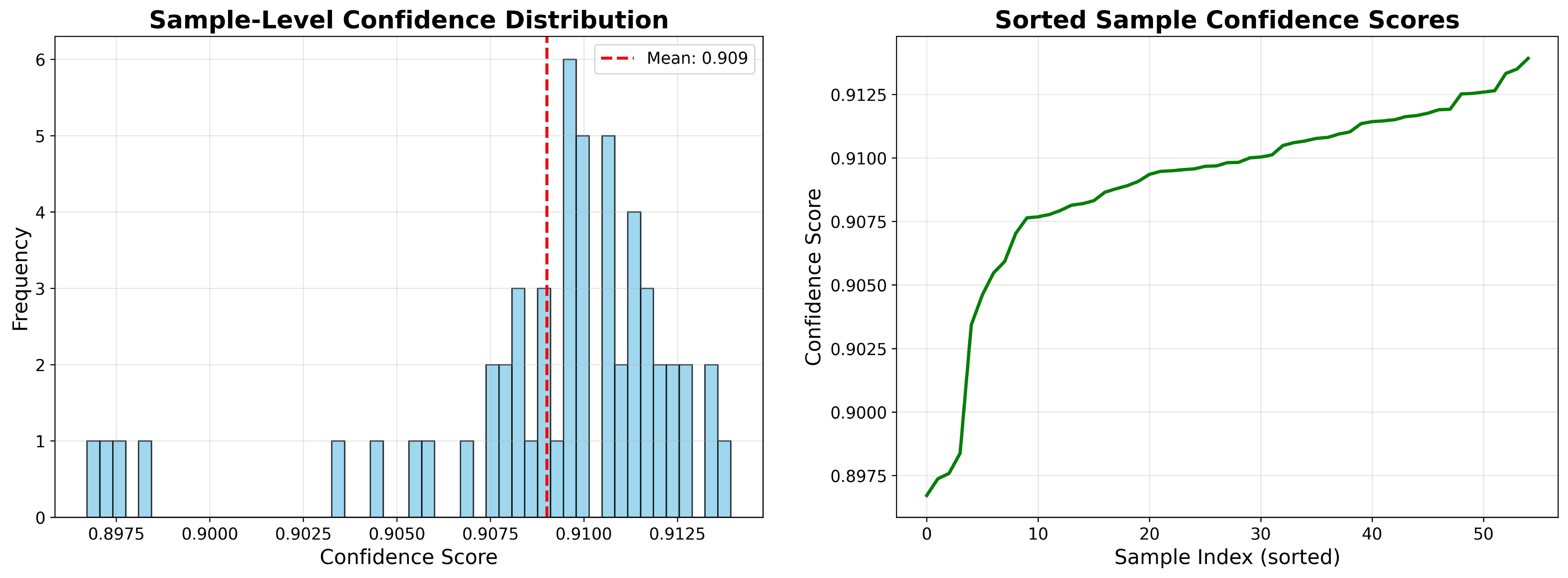}
  \caption{Statistical analysis of sample-level confidence scores for pseudo-labeled predictions.  Left: Histogram showing tight distribution around mean 0.909 (range: 0.8975-0.9125), indicating narrow variance and high overall confidence.  Right: Sorted confidence scores displaying smooth monotonic increase without outliers, validating the confidence-based ranking strategy for progressive sample selection.}
  \label{fig:confidence}
\end{figure}

To evaluate the confidence distribution of the model across individual samples, we conducted a statistical analysis of sample-level confidence scores. Figure \ref{fig:confidence} summarizes sample–level confidence.

The figure consists of two plots that illustrate the distribution and range of sample-level confidence scores. The histogram on the left shows the frequency distribution of confidence scores, which are tightly clustered around a mean value of 0.909 (marked by a red dashed line). The scores range from approximately 0.8975 to 0.9125, with the highest frequency occurring near the mean, indicating a relatively narrow spread. The plot on the right displays the sorted confidence scores, showing a steady increase from the minimum score of around 0.8975 to the maximum score of about 0.9125. The smooth curve suggests a consistent and gradual variation in the confidence scores across samples. Together, the plots provide a comprehensive view of the central tendency, spread, and variability of the confidence scores.

Performance Metrics of Student Model During Progressive Training Across Six Stages

Table~\ref{tab:training_stages} summarizes the performance evolution across six 
progressive training stages with increasing amounts of training data.

\begin{table*}[t]
\centering
\caption{Performance evolution across six progressive training stages with incremental pseudo-labeled data (10\%-100\%). Validation loss decreases from 1.789 to 0.878 and Dice coefficient improves from 0.393 to 0.872, with largest gains in early stagesand diminishing returns later.}
\label{tab:training_stages}
\begin{tabular}{lccccc}
\toprule
\textbf{Training Stage} & \textbf{\% of Data (K\%)} & \textbf{Best Val Loss} & \textbf{Best Val Dice} & \textbf{Dice Gain} & \textbf{Total Samples Used} \\
\midrule
Stage 1 & 10\%  & 1.789 & 0.393 & +0.1543 & $\sim$20\% \\
Stage 2 & 20\%  & 1.427 & 0.547 & +0.1543 & $\sim$35\% \\
Stage 3 & 40\%  & 1.229 & 0.655 & +0.1080 & $\sim$50\% \\
Stage 4 & 60\%  & 1.011 & 0.801 & +0.1454 & $\sim$65\% \\
Stage 5 & 80\%  & 0.933 & 0.846 & +0.0450 & $\sim$80\% \\
Stage 6 & 100\% & 0.878 & 0.872 & +0.0263 & 100\% \\
\bottomrule
\end{tabular}
\end{table*}

\noindent
The results show a steady decrease in validation loss (from 1.789 at 10\% data to 0.878 at  100\% data) and a consistent increase in Dice coefficient (from 0.393 to 0.872). The most significant improvements occur between Stages 1--2 and 3--4, while gains stabilize in later stages, indicating diminishing returns as more data is added. These experiments are designed to evaluate the impact of training set size on 
model performance and to demonstrate that incorporating more data yields 
substantial benefits.

\begin{figure}[htbp]
  \centering
  \includegraphics[width=\linewidth]{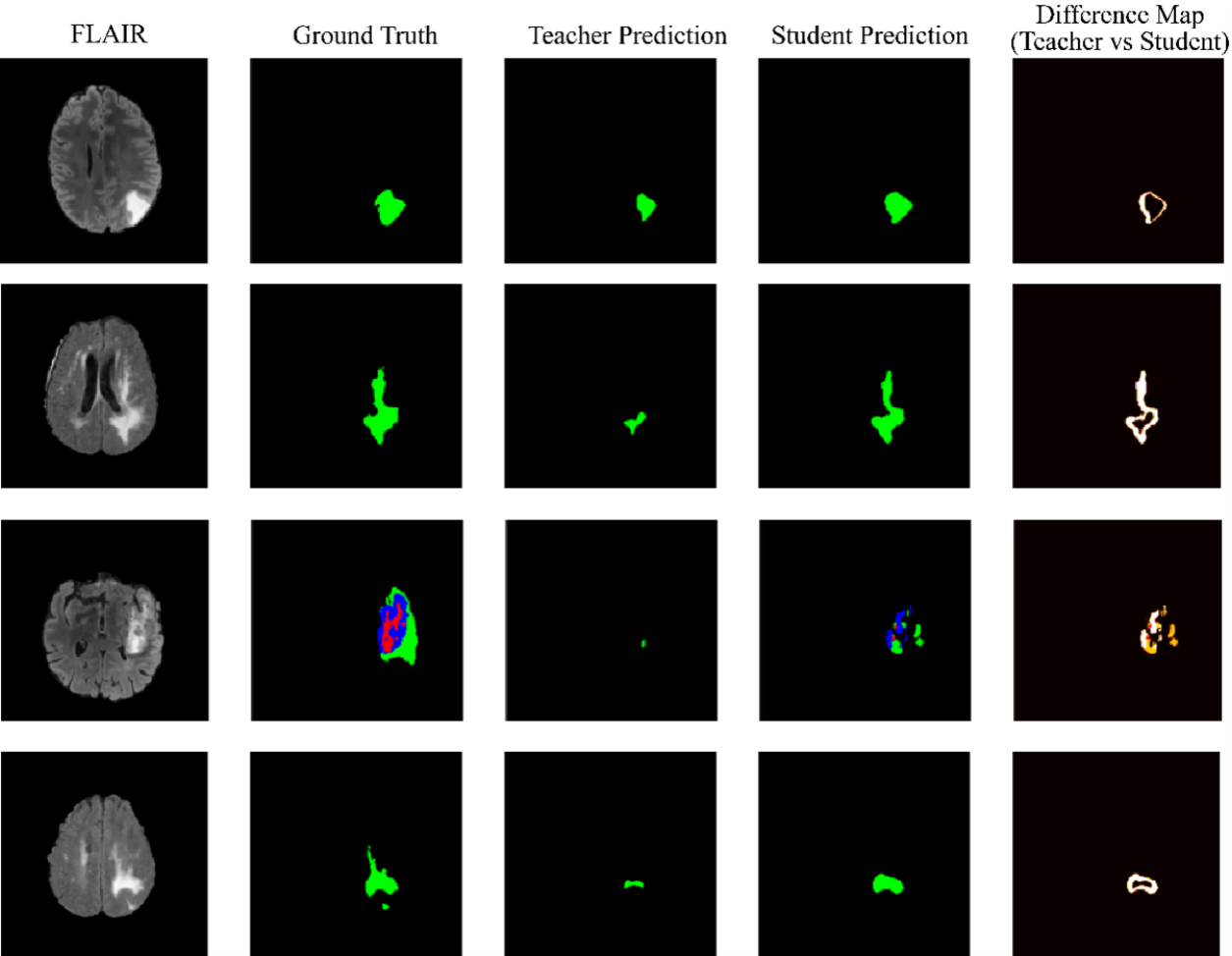}
  \caption{Qualitative comparison of teacher and student predictions on four FLAIR MRI cases.  Columns show: input image, ground truth, teacher prediction, student prediction, and difference map.  Both models accurately identify lesion cores with strong ground truth agreement.  Difference maps reveal subtle boundary variations and small structural discrepancies at lesion peripheries, validating effective knowledge transfer.}
  \label{fig:knowledge_transfer_flair}
\end{figure}

Figure~\ref{fig:knowledge_transfer_flair} presents qualitative results comparing the segmentation outputs of a teacher model and its corresponding student model on FLAIR MRI slices. Each row shows one example case, arranged in the columns.  

Figure~\ref{fig:knowledge_transfer_flair} provides a visual comparison between the segmentation outputs of a teacher model and a student model on FLAIR MRI scans, highlighting their alignment and differences. Each row corresponds to a different subject, with columns showing the FLAIR image, ground truth segmentation, teacher prediction, student prediction, and a difference map that captures the spatial discrepancies between teacher and student outputs. Both models generally show good agreement with the ground truth, accurately identifying lesion regions. However, subtle variations primarily around lesion boundaries and in smaller structures—are captured in the difference maps, where non-overlapping areas are outlined. These visualizations effectively demonstrate the fidelity of the student model in replicating the teacher’s behavior, offering qualitative insight into the knowledge transfer process and segmentation consistency across models.

\begin{figure}[htbp]
  \centering
  \includegraphics[width=\linewidth]{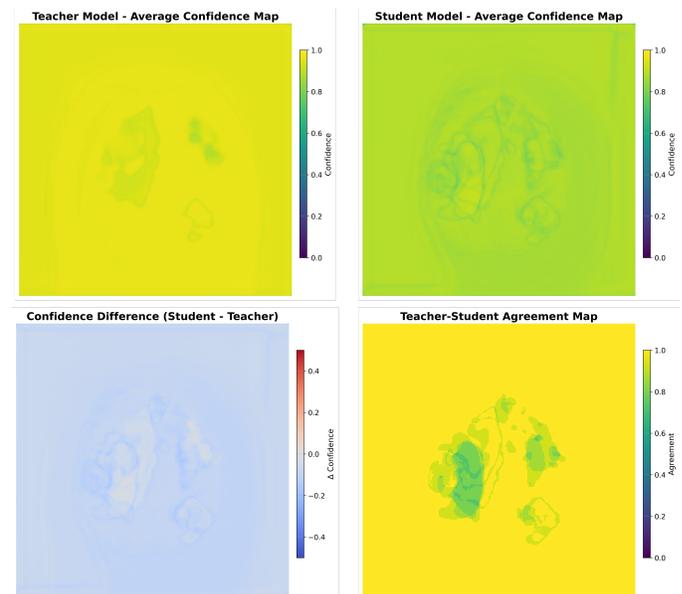}
  \caption{Spatial confidence and agreement analysis between teacher and student models.  Top: Pixel-wise confidence maps showing both models achieving high confidence (yellow regions) in lesion cores, with student exhibiting more boundary variation.  Bottom: Agreement map (left) and histogram (right) demonstrating strong prediction consensus concentrated at high agreement values ($>$0.9).}
  \label{fig:segmentation_model_confidence_analysis}
\end{figure}

Figure~\ref{fig:segmentation_model_confidence_analysis} presents a spatial analysis of prediction quality by comparing confidence and agreement maps between a teacher and student segmentation model. The top row displays the average confidence maps for the teacher (left) and student (right) models across a set of samples. Both models exhibit high overall confidence (closer to 1.0), with clearer structural detail in lesion areas, but the student model shows slightly more localized variation, suggesting heightened sensitivity to lesion boundaries.

\par The table \ref{tab:teacher_student_performance_summary} presents a comprehensive overview of the teacher-student learning framework applied to medical image segmentation. It consolidates key performance metrics across class-wise Dice and IoU scores, progressive learning stages, confidence analysis, and final outcomes. Notably, the student model outperforms the teacher in tumor-specific classes such as NCR/NET, Edema, and Enhancing, despite the teacher achieving a slightly higher overall Dice score. Progressive learning results show a steady improvement in the student’s validation Dice coefficient across six stages, with the most significant gains occurring in early stages as more unlabeled data is incrementally introduced. Confidence vs. quality analysis reveals a positive correlation, validating confidence as a proxy for prediction quality. The distribution of pseudo-label confidence further emphasizes that high-confidence predictions are concentrated in later stages.

\begin{table*}[t]
\centering
\caption{Unified summary of teacher-student framework performance and learning dynamics.}
\begin{tblr}{
  cells = {c},
  cell{2}{1} = {r=4}{},
  cell{6}{1} = {r=4}{},
  cell{10}{1} = {r=6}{},
  cell{17}{1} = {r=3}{},
  hline{1,23} = {-}{0.12em},
  hline{2,6,10,16-17,20-22} = {-}{0.05em},
}
\textbf{Metric}            & \textbf{Subcategory} & \textbf{Teacher} & \textbf{Student}           & \textbf{Notes}                           \\
Class-wise Dice            & Background           & 0.992            & 0.995                      & Strong agreement                         \\
                           & NCR/NET              & 0.731            & 0.797                      & Student improves on low-signal class     \\
                           & Edema                & 0.767            & 0.980                      & Substantial student improvement          \\
                           & Enhancing            & 0.000            & 0.620                      & Teacher fails; Student recovers          \\
Class-wise IoU             & Background           & 0.984            & 0.990                      & High IoU                                 \\
                           & NCR/NET              & 0.000            & 0.795                      & Student gains IoU where teacher has none \\
                           & Edema                & 0.691            & 0.915                      & Substantial student improvement          \\
                           & Enhancing            & 0.000            & 0.791                      & Teacher fails; student recovers          \\
Progressive Dice per Stage & Stage 1 (10\%)       & -                & 0.393                      & Base performance                         \\
                           & Stage 2 (20\%)       & -                & 0.547                      & $\bigtriangleup$ +0.154                  \\
                           & Stage 3 (40\%)       & -                & 0.655                      & $\bigtriangleup$ +0.108                  \\
                           & Stage 4 (60\%)       & -                & 0.801                      & $\bigtriangleup$ +0.145                  \\
                           & Stage 5 (80\%)       & -                & 0.846                      & $\bigtriangleup$ +0.045                  \\
                           & Stage 6 (100\%)      & -                & 0.872                      & $\bigtriangleup$ +0.026                  \\
Confidence vs Quality      & 0.853–0.860          & -                & $\uparrow$ Quality         & Positive correlation                     \\
Pseudo-label Confidence    & Top 10\%             & -                & Peak at $\sim$ 0.860-0.861 & High confidence concentration            \\
                           & Top 20\%             & -                & Broader spread             & High confidence concentration            \\
                           & Top 40\%             & -                & Even wider                 & High confidence concentration            \\
Final Dice Score           & Best Value           & 0.8217           & 0.8721                     & $\bigtriangleup$ 0.0504                 \\
Final Accuracy             & Overall              & -                & 0.9881                     & -                                        \\
Training Stages            & -                    & -                & 6                          & -                                        
\end{tblr}
\label{tab:teacher_student_performance_summary}
\end{table*}

\begin{figure}[htbp]
  \centering
  \includegraphics[width=\linewidth]{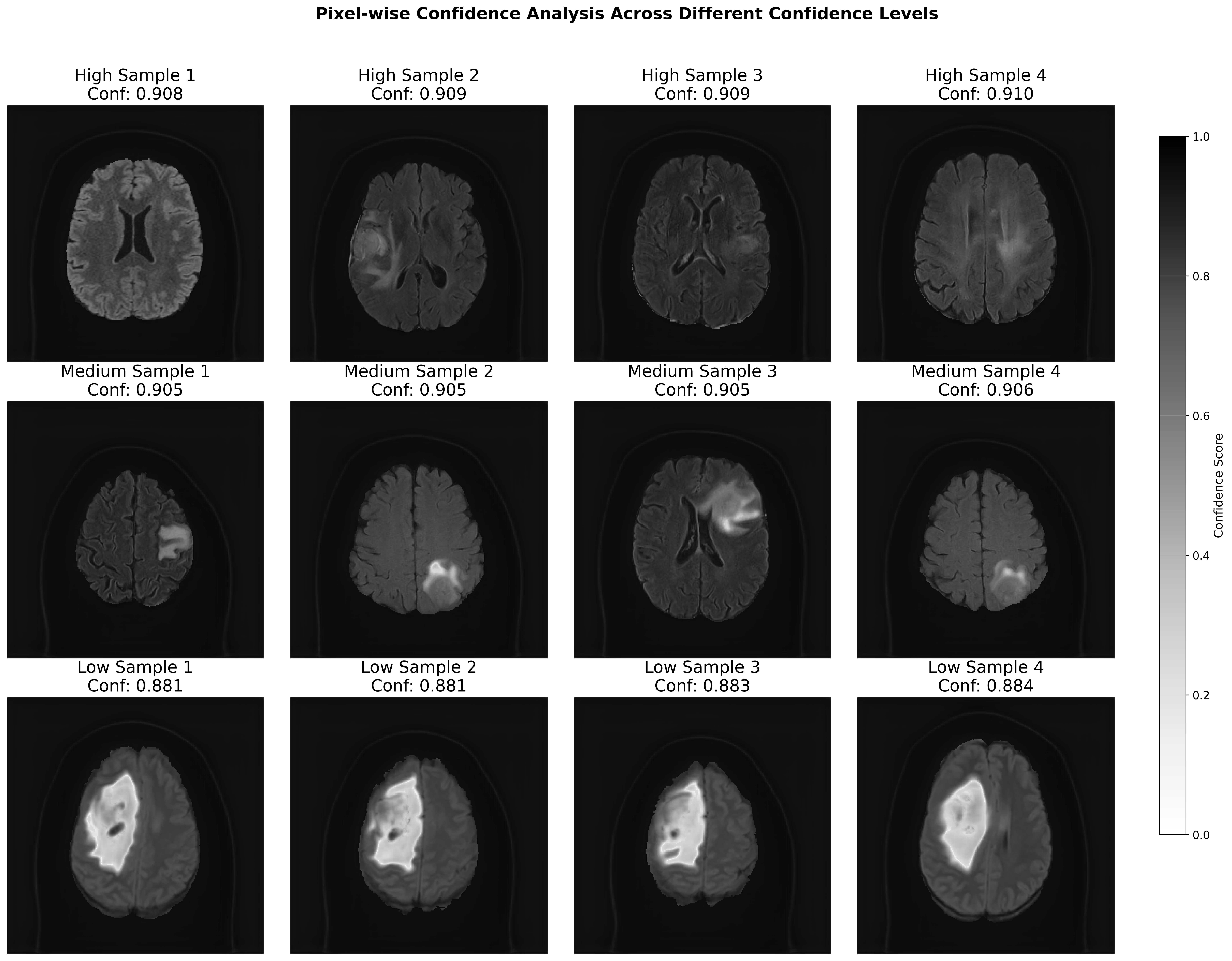}
   \caption{Pixel-wise confidence stratification across brain MRI samples categorized by prediction confidence levels. Three rows display high (0.908-0.910), medium (0.905-0.906), and low (0.881-0.884) confidence predictions, each with four representative slices and annotated mean scores. Grayscale colorbar indicates confidence intensity. High-confidence samples exhibit uniform lesion boundaries, while low-confidence samples show heterogeneous patterns, demonstrating spatially-resolved uncertainty estimation for selective prediction use.}
  \label{fig:pixel_wise_confidence_analysis}
\end{figure}

Figure \ref{fig:pixel_wise_confidence_analysis} illustrates pixel-wise prediction confidence maps for a set of segmentation outputs, stratified by overall confidence levels—high, medium, and low. Each row contains four representative MRI slices from each confidence tier, with the corresponding average confidence score annotated above each sample. 

The figure\ref{fig:pixel_wise_confidence_analysis} presents a grid of brain MRI scans arranged to analyze pixel-wise confidence levels across different samples with varying confidence scores. The rows represent samples categorized into high, medium, and low confidence levels, with corresponding confidence values noted beneath each image. The high-confidence samples (first row) exhibit confidence scores ranging from 0.908 to 0.910, while the medium-confidence samples (second row) range from 0.905 to 0.906, and the low-confidence samples (third row) vary from 0.881 to 0.884. A grayscale color bar on the right indicates the confidence score, with lighter regions in the scans suggesting areas of higher confidence. The visualization highlights the spatial distribution of confidence scores and how these vary across different samples, providing insights into the pixel-level confidence variability within the MRI images. The overall layout emphasizes differences between confidence levels while maintaining consistency in the display format for comparison.

\begin{figure}[htbp]
  \centering
  \includegraphics[width=\linewidth]{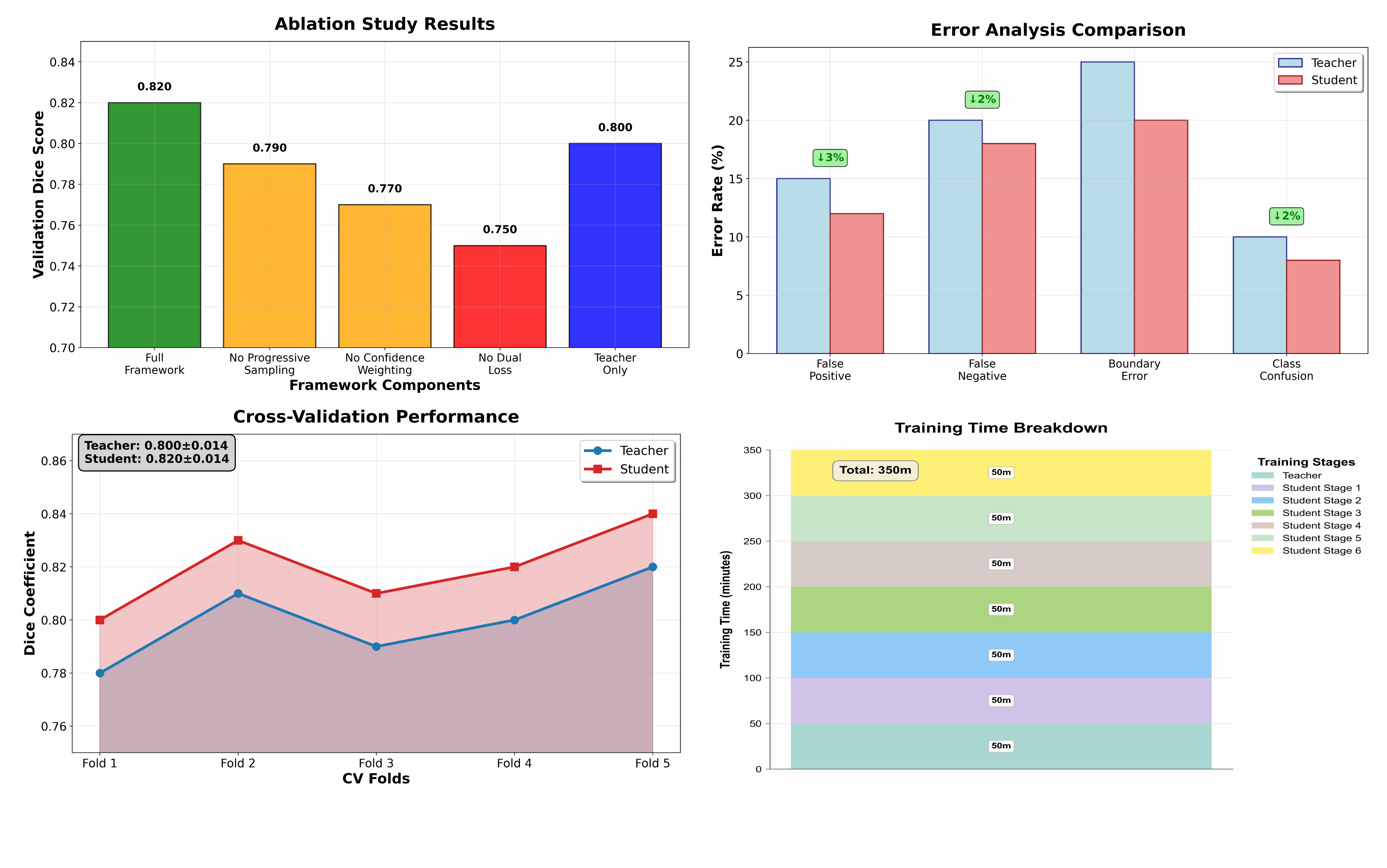}
  \caption{Comprehensive performance analysis of the teacher-student framework across four dimensions. Top-left: Ablation study showing full framework achieves highest Dice (0.820) versus reduced variants (0.785-0.801). Top-right: Error analysis indicating student reduces false positives (-3\%) and class confusion (-2\%) but increases boundary errors (+5\%). Bottom-left: Five-fold cross-validation demonstrating student (0.872$\pm$ 0.014) outperforming teacher (0.821$\pm$ 0.014). Bottom-right: Training time totaling 350 minutes distributed evenly across teacher and six student stages.}
  \label{fig:comprehensive_performance_analysis}
\end{figure}

Finally, we present a comprehensive performance analysis for a teacher-student framework designed for segmentation tasks. It encapsulates multiple facets of the training process, providing a holistic view of the framework's effectiveness, efficiency, and potential directions for enhancement. Figure \ref{fig:comprehensive_performance_analysis} provides a detailed evaluation of the teacher-student framework applied to image segmentation tasks using ablation study results, error analysis, cross-validation performance, and training time breakdown.

This figure presents a comprehensive evaluation of this framework through four subplots. The Ablation Study Results (top left) demonstrate that the full framework achieves the highest validation Dice score (0.820), while removing components like progressive sampling, confidence weighting, or dual loss reduces performance. The Error Analysis Comparison (top right) shows that the student model improves error rates for false positives (-3\%) and class confusion (-2\%) but performs slightly worse in false negatives (-2\%) and boundary errors (-5\%) compared to the teacher model. The Cross-Validation Performance (bottom left) indicates that the Student model slightly outperforms the Teacher model in terms of average Dice coefficient (0.872 ± 0.014 vs. 0.821 ± 0.014) across five folds. Finally, the Training Time Breakdown (bottom right) highlights a total training time of 350 minutes, evenly distributed across the Teacher model and six stages of the Student model, each taking 50 minutes. Overall, the student model demonstrates better or comparable performance to the teacher model while maintaining efficiency.

\section{Discussion}

\subsection{Technical Performance and Clinical Relevance}

The proposed semi-supervised teacher-student framework integrates uncertainty-aware pseudo-labeling with progressive confidence-based sampling to address critical challenges in automated brain tumor segmentation. The teacher network generates pseudo-labels with associated confidence estimates, while the student model optimizes a dual-loss objective that promotes learning from high-reliability regions while suppressing overfitting to uncertain predictions. Quantitative evaluation demonstrates Dice Similarity Coefficient scores of 0.797 for necrotic core and non-enhancing tumor (NCR/NET), 0.980 for peritumoral edema, and 0.620 for enhancing tumor components, reflecting enhanced capacity to capture subtle structural boundaries and intensity heterogeneity across tumor subregions.

These technical achievements translate directly into clinical utility. The enhancing tumor component, characterized by contrast uptake on T1-weighted post-gadolinium sequences, serves as a radiological biomarker of blood-brain barrier disruption and proliferative tumor activity. Precise identification of enhancing margins (DSC 0.620) is essential for defining surgical resection boundaries and stereotactic radiosurgery target volumes. The exceptional edema segmentation accuracy (DSC 0.980) enables quantitative assessment of vasogenic edema extent, directly informing corticosteroid dosing protocols and surgical urgency decisions. By automating multi-sequence MRI annotation, this framework reduces radiologist workload, standardizes volumetric measurements, and enables clinicians to focus on complex surgical strategy optimization.

\subsection{Addressing Data Scarcity and Multi-Institutional Deployment}

A critical barrier to clinical translation is the limited availability of expert-annotated datasets. Manual glioma segmentation across four MRI modalities demands specialized neuroradiological expertise and substantial time investment. This semi-supervised paradigm achieves competitive performance using only 10-20\% labeled data, with unlabeled scans progressively incorporated through confidence-guided pseudo-labeling. This data efficiency democratizes access to AI-assisted diagnostic tools across heterogeneous clinical environments, from academic medical centers to community hospitals with limited subspecialty resources.

The framework's generalizability across imaging protocols and scanner types addresses inter-institutional variability—a persistent challenge in neuro-oncology. Glioma patients commonly encounter variations in field strength (1.5T versus 3T), pulse sequences, and scanning protocols across facilities. Supervised models trained on single-center data often show performance degradation on external datasets, limiting utility in multi-center networks. By incorporating unlabeled data from diverse sources, our approach learns tumor-specific features rather than site-specific artifacts, enabling standardized segmentation tools where reliable volumetric measurements are essential for treatment response evaluation and clinical research. The confidence-based quality metrics provide interpretable uncertainty quantification for automatic case flagging and expert review—critical for regulatory compliance and medico-legal risk management. These capabilities position the framework as a practical solution for longitudinal disease monitoring, where serial MRI examinations require consistent volumetric analysis throughout treatment courses.

\section{Conclusion}

We introduced a semi-supervised teacher–student framework that couples uncertainty-aware pseudo-labels with a distribution-driven, progressive sampling curriculum and a dual-loss objective that discourages learning from low-confidence regions. Experiments on BraTS 2021 show consistent loss reductions and DSC improvements across stages, with strong early gains and stable convergence. While the teacher attains higher global DSC, the student model improves clinically relevant tumor subregions and successfully handles cases where the teacher under-segments.

Future work can be focused on external multi-site validation and dynamic threshold scheduling. moreover, integrating active learning and federated training can be experimented with to further reduce annotation needs and enhance deployment in real clinical workflows.


\begin{thebibliography}{1}

\bibitem{hosny2018artificial}
A.~Hosny, C.~Parmar, J.~Quackenbush, L.~H.~Schwartz, and H.~J.~W.~L.~Aerts, ``Artificial intelligence in radiology,'' \emph{Nat. Rev. Cancer}, vol.~18, no.~8, pp.~500--510, Aug. 2018.


\bibitem{esteva2019guide}
A.~Esteva \emph{et al.}, ``A guide to deep learning in healthcare,'' \emph{Nat. Med.}, vol.~25, no.~1, pp.~24--29, Jan. 2019.

\bibitem{ronneberger2015unet}
O.~Ronneberger, P.~Fischer, and T.~Brox, ``U-net: Convolutional networks for
  biomedical image segmentation,'' in \emph{Medical Image Computing and
  Computer-Assisted Intervention (MICCAI)}, Munich, Germany, 2015, pp.
  234--241.

\bibitem{karun2025hybrid}
B.~Karun, A.~Thiyagarajan, P.~R. Murugan, N.~Jeyaprakash, K.~Ramaraj, and
  R.~Makreri, ``Advanced hybrid brain tumor segmentation in {MRI}: Elephant
  herding optimization combined with entropy-guided fuzzy clustering,''
  \emph{Math. Comput. Appl.}, vol.~30, no.~1, p.~1, 2025.

\bibitem{bai2017semi}
W.~Bai, O.~Oktay, M.~Sinclair, M.~Rajchl, G.~Tarroni, B.~Glocker, and A.~P.
  King, ``Semi-supervised learning for network-based cardiac {MR} image
  segmentation,'' in \emph{Medical Image Computing and Computer-Assisted
  Intervention (MICCAI)}, 2017, pp. 253--260.

\bibitem{huang2023selfsupervised_review}
S.~C. Huang, A.~Pareek, M.~Jensen, \emph{et~al.}, ``Self-supervised learning
  for medical image classification: a systematic review and implementation
  guidelines,'' \emph{npj Digit. Med.}, vol.~6, p.~74, 2023.

\bibitem{wang2023ssl}
Z.~Wang, L.~Zhou, Y.~He, S.~Li, and D.~Shen, ``Semi-supervised {3D} brain tumor
  segmentation using mean teacher with uncertainty-aware pseudo-labels,''
  \emph{Med. Image Anal.}, vol.~87, p. 102799, 2023.

\bibitem{yu2019uncertainty}
L.~Yu, S.~Wang, X.~Li, C.-W. Fu, and P.-A. Heng, ``Uncertainty-aware
  self-ensembling model for semi-supervised {3D} left atrium segmentation,'' in
  \emph{Int. Conf. Med. Image Comput. Comput.-Assisted Intervent.}, 2019, pp.
  605--613.

\bibitem{tarvainen2017mean}
A.~Tarvainen and H.~Valpola, ``Mean teachers are better role models:
  Weight-averaged consistency targets improve semi-supervised deep learning
  results,'' 2017, arXiv:1703.01780. [Online]. Available:
  https://arxiv.org/abs/1703.01780

\bibitem{li2020sassnet}
S.~Li, B.~Liu, L.~Zhang, S.~Cui, Z.~Liu, N.~Dong, and X.~Zhu, ``Shape-aware
  semi-supervised {3D} semantic segmentation for medical images,'' in
  \emph{Proc. Int. Conf. Med. Image Comput. Comput.-Assisted Intervent.
  (MICCAI)}, Lima, Peru, Oct. 2020, pp. 552--561.

\bibitem{chen2021adversarialmt}
G.~Chen, J.~Ru, Y.~Zhou, I.~Rekik, Z.~Pan, X.~Liu, Y.~Lin, B.~Lu, and J.~Shi,
  ``{MTANS}: Multi-scale mean teacher combined adversarial network with
  shape-aware embedding for semi-supervised brain lesion segmentation,''
  \emph{NeuroImage}, vol. 244, p. 118568, 2021.

\bibitem{xu2023ambiguity}
Z.~Xu, Y.~Wang, D.~Lu, X.~Luo, J.~Yan, Y.~Zheng, and R.~K.-y. Tong,
  ``Ambiguity-selective consistency regularization for mean-teacher
  semi-supervised medical image segmentation,'' \emph{Med. Image Anal.}, vol.~88,
  p. 102880, 2023.

\bibitem{lu2023uncertainty}
L.~Lu, M.~Yin, L.~Fu, and F.~Yang, ``Uncertainty-aware pseudo-label and
  consistency for semi-supervised medical image segmentation,'' \emph{Biomed.
  Signal Process. Control}, vol.~79, p. 104203, 2023.

\bibitem{Chen2021CPS}
X.~Chen, Y.~Yuan, G.~Zeng, and J.~Wang, ``Semi-supervised semantic segmentation
  with cross pseudo supervision,'' in \emph{Proc. IEEE/CVF Conf. Comput. Vis.
  Pattern Recognit. (CVPR)}, Nashville, TN, USA, Jun. 2021, pp. 2613--2622.

\bibitem{wu2022mcnet}
Y.~Wu, Z.~Ge, D.~Zhang, M.~Xu, L.~Zhang, Y.~Xia, and J.~Cai, ``Mutual
  consistency learning for semi-supervised medical image segmentation,''
  \emph{Med. Image Anal.}, vol.~81, p. 102530, 2022.

\bibitem{Liu2024MOST}
X.~Liu, Z.~Chen, and Y.~Yuan, ``{MOST}: Multi-formation soft masking for
  semi-supervised medical image segmentation,'' in \emph{Proc. Int. Conf. Med.
  Image Comput. Comput.-Assisted Intervent. (MICCAI)}, Marrakesh, Morocco, Oct.
  2024, pp. 469--480.

\bibitem{zhu2024dualmt}
J.~Zhu, B.~Bolsterlee, B.~V.~Y. Chow, Y.~Song, and E.~Meijering, ``Hybrid dual
  mean-teacher network with double-uncertainty guidance for semi-supervised
  segmentation of magnetic resonance images,'' \emph{Comput. Med. Imaging
  Graph.}, vol. 115, p. 102383, Jul. 2024.

\bibitem{He2025EPL}
Y.~He, L.~Li, T.~Zhan, C.-M. Pun, W.~Jiao, and Z.~Jin, ``Evidential prototype
  learning for semi-supervised medical image segmentation,'' in \emph{Proc.
  31st ACM SIGKDD Conf. Knowl. Discov. Data Mining (KDD)}, Toronto, ON, Canada,
  2025, pp. 908--919.

\bibitem{assefa2025dycon}
M.~Assefa, M.~Naseer, I.~I. Ganapathi, \emph{et~al.}, ``{DyCON}: Dynamic
  uncertainty-aware consistency and contrastive learning for semi-supervised
  medical image segmentation,'' in \emph{CVPR}, 2025.

\bibitem{karri2024ugcmt}
M.~Karri, A.~S. Arya, K.~Biswas, \emph{et~al.}, ``Uncertainty-guided cross
  attention ensemble mean teacher for semi-supervised medical image
  segmentation,'' in \emph{Proc. IEEE/CVF Winter Conf. Appl. Comput. Vis.
  (WACV)}, Tucson, AZ, USA, 2025, pp. 7039--7048.

\bibitem{Qu_2024}
G.~Qu, B.~Lu, J.~Shi, Z.~Wang, Y.~Yuan, Y.~Xia, Z.~Pan, and Y.~Lin,
  ``Motion-artifact-augmented pseudo-label network for semi-supervised brain
  tumor segmentation,'' \emph{Phys. Med. Biol.}, vol.~69, no.~5, p. 055023,
  Feb. 2024.

\bibitem{ottesen2025ssl}
J.~A. Ottesen, E.~Tong, K.~E. Emblem, A.~Latysheva, G.~Zaharchuk,
  A.~Bj{\o}rnerud, and E.~Gr{\o}vik, ``Semi-supervised learning allows for
  improved segmentation with reduced annotations of brain metastases using
  multicenter {MRI} data,'' \emph{J. Magn. Reson. Imaging}, vol.~61, no.~6, pp.
  2469--2479, 2025.

\bibitem{menze2015brats}
B.~H. Menze, A.~Jakab, S.~Bauer, J.~Kalpathy-Cramer, K.~Farahani, J.~Kirby,
  \emph{et~al.}, ``The multimodal brain tumor image segmentation benchmark
  ({BRATS}),'' \emph{IEEE Trans. Med. Imaging}, vol.~34, no.~10, pp.
  1993--2024, 2015.

\bibitem{Lakshmi2025}
K.~Lakshmi, S.~Amaran, G.~Subbulakshmi, \emph{et~al.}, ``Explainable artificial
  intelligence with {UNet}-based segmentation and {Bayesian} machine learning
  for classification of brain tumors using {MRI} images,'' \emph{Sci. Rep.},
  vol.~15, p. 690, 2025.

\bibitem{Emon2025}
S.~H. Emon, T.-L.~B. Tseng, M.~Pokojovy, S.~Moen, P.~McCaffrey, E.~Walser,
  A.~Vo, and M.~F. Rahman, ``Uncertainty-guided semi-supervised ({UGSS}) mean
  teacher framework for brain hemorrhage segmentation and volume
  quantification,'' \emph{Biomed. Signal Process. Control}, vol. 102, p.
  107386, 2025.

\bibitem{Wang2024}
P.~Wang, Y.~Liu, and Z.~Zhou, ``Supraspinatus extraction from {MRI} based on
  attention-dense spatial pyramid {UNet} network,'' \emph{J. Orthop. Surg.
  Res.}, vol.~19, p.~60, 2024.

\bibitem{LIU2024102776}
H.~Liu, J.~Huang, Q.~Li, X.~Guan, and M.~Tseng, ``A deep convolutional neural
  network for the automatic segmentation of glioblastoma brain tumor: Joint
  spatial pyramid module and attention mechanism network,'' \emph{Artif. Intell.
  Med.}, vol. 148, p. 102776, 2024.

\bibitem{HANINE2024109440}
L.~Hanine and N.~Hilal, ``Hybrid depthwise convolution bottleneck in a {UNet}
  architecture for advanced brain tumor segmentation,'' \emph{Eng. Appl. Artif.
  Intell.}, vol. 138, p. 109440, 2024.

\bibitem{Panigrahi2024}
S.~Panigrahi, D.~R.~D. Adhikary, and B.~K. Pattanayak, ``Analyzing activation
  functions with transfer learning-based layer customization for improved brain
  tumor classification,'' \emph{IEEE Access}, vol.~12, pp. 16\,870--16\,891,
  2024.

\bibitem{Wang2024ICBASE}
Z.~Wang, Y.~Chen, F.~Wang, and Q.~Bao, ``Improved {UNet} model for brain tumor
  image segmentation based on {ASPP}-coordinate attention mechanism,'' in
  \emph{2024 5th Int. Conf. Big Data \& Artif. Intell. \& Softw. Eng.
  (ICBASE)}, Wenzhou, China, 2024, pp. 393--397.

\bibitem{Rasool2024}
N.~Rasool, J.~I. Bhat, N.~A. Wani, N.~Ahmad, and M.~Alshara, ``{TransResUNet}:
  Revolutionizing glioma brain tumor segmentation through transformer-enhanced
  residual {UNet},'' \emph{IEEE Access}, vol.~12, pp. 72\,105--72\,116, 2024.

\bibitem{AlHasan2025}
S.~Al~Hasan, S.~M. Mahim, M.~E. Hossen, \emph{et~al.}, ``{DSIT UNet}: a dual
  stream iterative transformer-based {UNet} architecture for segmenting brain
  tumors from {FLAIR MRI} images,'' \emph{Sci. Rep.}, vol.~15, p. 13815, 2025.

\bibitem{ZHU2024108284}
Z.~Zhu, M.~Sun, G.~Qi, Y.~Li, X.~Gao, and Y.~Liu, ``Sparse dynamic volume
  {TransUNet} with multi-level edge fusion for brain tumor segmentation,''
  \emph{Comput. Biol. Med.}, vol. 172, p. 108284, 2024.

\bibitem{Tiwary2025}
P.~K. Tiwary, P.~Johri, A.~Katiyar, and M.~K. Chhipa, ``Deep learning-based
  {MRI} brain tumor segmentation with {EfficientNet}-enhanced {UNet},''
  \emph{IEEE Access}, vol.~13, pp. 54\,920--54\,937, 2025.

\bibitem{Hasan2024}
S.~A. Hasan, S.~M. Mahim, M.~E. Hossen, M.~O. Hasan, T.~H. Ashik, F.~Ahmmed,
  M.~K. Islam, and M.~S. Miah, ``{DSP-UNet}: Dual-skip perceiver {UNet} for
  lower-grade glioma segmentation,'' in \emph{2024 27th Int. Conf. Comput. Inf.
  Technol. (ICCIT)}, Cox's Bazar, Bangladesh, 2024, pp. 3069--3074.

\bibitem{AN2024110731}
D.~An, P.~Liu, Y.~Feng, P.~Ding, W.~Zhou, and B.~Yu, ``Dynamic weighted
  knowledge distillation for brain tumor segmentation,'' \emph{Pattern
  Recognit.}, vol. 155, p. 110731, 2024.

\bibitem{Guo2025}
X.~Guo \emph{et~al.}, ``Uncertainty driven adaptive self-knowledge distillation
  for medical image segmentation,'' \emph{IEEE Trans. Emerg. Topics Comput.
  Intell.}, vol.~9, no.~5, pp. 3455--3468, 2025.

\bibitem{Rasool2025Review}
N.~Rasool and J.~I. Bhat, ``A critical review on segmentation of glioma brain
  tumor and prediction of overall survival,'' \emph{Arch. Comput. Methods
  Eng.}, vol.~32, pp. 1525--1569, 2025.

\bibitem{Hassan2024}
R.~Hassan, M.~R.~H. Mondal, and S.~I. Ahamed, ``{UDBRNet}: A novel uncertainty
  driven boundary refined network for organ at risk segmentation,'' \emph{PLOS
  ONE}, vol.~19, no.~6, pp. 1--18, Jun. 2024.

\bibitem{Yang2024HippocampusSegmentation}
Q.~Yang, C.~Wang, K.~Pan, \emph{et~al.}, ``An improved {3D-UNet}-based brain
  hippocampus segmentation model based on {MR} images,'' \emph{BMC Med.
  Imaging}, vol.~24, no.~1, p. 166, 2024.

\bibitem{Chatterjee2025PseudoLabel}
P.~Chatterjee, K.~D. Sharma, and A.~Chakrabarti, ``Refining pseudo labels with
  a teacher-student paradigm for accurate brain lesion segmentation,''
  \emph{IEEE Access}, vol.~13, pp. 84\,474--84\,487, 2025.

\bibitem{Zhang_2024_CVPR}
X.~Zhang, Y.~Wu, E.~Angelini, A.~Li, J.~Guo, J.~M. Rasmussen, T.~G. O'Connor,
  P.~D. Wadhwa, A.~P. Jackowski, H.~Li, J.~Posner, A.~F. Laine, and Y.~Wang,
  ``{MAPSeg}: Unified unsupervised domain adaptation for heterogeneous medical
  image segmentation based on {3D} masked autoencoding and pseudo-labeling,''
  in \emph{Proc. IEEE/CVF Conf. Comput. Vis. Pattern Recognit. (CVPR)}, Jun.
  2024, pp. 5851--5862.

\bibitem{Ran2025PseudoLabelReview}
L.~Ran, Y.~Li, G.~Liang, and Y.~Zhang, ``Pseudo labeling methods for
  semi-supervised semantic segmentation: A review and future perspectives,''
  \emph{IEEE Trans. Circuits Syst. Video Technol.}, vol.~35, no.~4, pp.
  3054--3080, 2025.

\bibitem{Zheng2025MTSNet}
Y.~Zheng, Z.~Sun, S.~Chen, \emph{et~al.}, ``{MTS-Net}: Research on multi-modal,
  multi-task, multi-stage tumor segmentation model for {PET/CT},'' \emph{Appl.
  Intell.}, vol.~55, p. 870, 2025.

\bibitem{AlZubi2025Federated}
T.~M. AlZubi and H.~Mukhtar, ``Federated knowledge distillation with {3D}
  transformer adaptation for weakly labeled multi-organ medical image
  segmentation,'' \emph{IEEE Access}, vol.~13, pp. 83\,619--83\,642, 2025.

\bibitem{Ziaee2025}
S.~Ziaee, ``Uncertainty-guided medical image segmentation,'' Master's thesis,
  University of Calgary, Calgary, Canada, 2025.

\bibitem{Arora2024KidneySegmentation}
R.~K. Arora, A.~Kumar, and A.~Soni, ``Deep learning approaches for enhanced
  kidney segmentation: Evaluating {U-Net} and attention {U-Net} with
  cross-entropy and focal loss functions,'' in \emph{2024 IEEE 3rd World Conf.
  Appl. Intell. Comput. (AIC)}, Gwalior, India, 2024, pp. 854--860.

\bibitem{AbouAli2025}
M.~A. Ali, J.~Charafeddine, F.~Dornaika, \emph{et~al.}, ``Enhancing
  generalization and mitigating overfitting in deep learning for brain cancer
  diagnosis from {MRI},'' \emph{Appl. Magn. Reson.}, vol.~56, pp. 359--394,
  2025.

\bibitem{PANI2024109418}
K.~Pani and I.~Chawla, ``A hybrid approach for multi-modal brain tumor
  segmentation using two-phase transfer learning, {SSL} and a hybrid {3D
  UNet},'' \emph{Comput. Electr. Eng.}, vol. 118, p. 109418, 2024.



\end{thebibliography}
\end{document}